\documentclass[10pt,twocolumn,letterpaper]{article}

\usepackage[pagenumbers]{cvpr} 

\usepackage{microtype}
\usepackage{xcolor}
\usepackage{xspace}
\usepackage{graphicx}
\usepackage{svg}
\usepackage{algorithm}
\usepackage{algorithmic}
\usepackage{subcaption}
\usepackage{booktabs}
\usepackage{array}
\usepackage{multirow}
\usepackage{afterpage}
\usepackage{amsmath}
\usepackage{amssymb}
\usepackage{mathtools}
\usepackage{amsthm}

\definecolor{cvprblue}{rgb}{0.21,0.49,0.74}
\usepackage[pagebackref,breaklinks,colorlinks,allcolors=cvprblue]{hyperref}
\usepackage[capitalize,noabbrev]{cleveref}

\def\paperID{15} 
\def\confName{CVPR}
\def\confYear{2026}

\title{Guidance for Low-Level Perceptual Editing in Unconditional Diffusion Models}

\author{
\begin{tabular}{ccc}
Shreyansh Modi\thanks{Equal contribution.}\thanks{Indian Institute of Technology Roorkee.} & Akshat Tomar\footnotemark[1]\footnotemark[2] & Aarush Aggarwal\footnotemark[1]\footnotemark[2]\\
{\tt\small shreyansh\_m@ee.iitr.ac.in} &
{\tt\small akshat\_t@mfs.iitr.ac.in} &
{\tt\small aarush\_a@ma.iitr.ac.in}
\end{tabular}
}

\begin{document}
\maketitle
\begin{abstract}
Unconditional diffusion models offer powerful generative priors, yet steering them toward aesthetically enhanced outputs remains largely unexplored. We show that h-space patching, the dominant paradigm for training-free diffusion editing, systematically fails for global, low-level transformations required for aesthetic and perceptual refinement. We introduce a novel, generalized framework for image-editing in unconditional diffusion models without explicit training. This inference-time mechanism operates on low-level features by extracting degradation concept vectors and combining bottleneck patching with classifier-free guidance to guide sampling away from the degraded manifold, producing consistently improved images without any model retraining.
\end{abstract}

\section{Introduction}
Diffusion models have emerged as the state-of-the-art paradigm for image synthesis~\citep{ho2020ddpm}, with deterministic DDIM sampling \citep{song2020ddim} enabling near-perfect reconstruction.

Guidance mechanisms \citep{dhariwal2021diffusion,ho2022cfg} further steer the reverse diffusion process toward desired distributions, with recent inference-time methods constructing negative baselines by intervening directly in internal representations \citep{hong2023sag,ahn2024pag,hong2024seg,shen2024scfg,fu2025layeredit} circumventing the need for conditional training entirely.

\begin{figure}[!t]
\centering
\setlength{\tabcolsep}{0pt}
\renewcommand{\arraystretch}{0.85}
\captionsetup{skip=1pt}
\begin{tabular}{@{}>{\centering\arraybackslash}m{0.075\columnwidth}@{\hspace{1.5pt}}>{\centering\arraybackslash}m{0.298\columnwidth}@{\hspace{1.5pt}}>{\centering\arraybackslash}m{0.298\columnwidth}@{\hspace{1.5pt}}>{\centering\arraybackslash}m{0.298\columnwidth}@{}}
{\rotatebox[origin=c]{90}{\footnotesize{Baseline}}} &
\includegraphics[width=\linewidth]{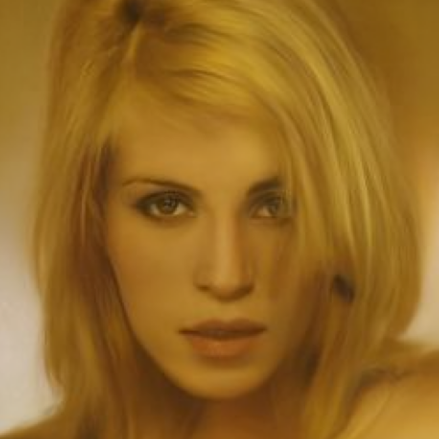} &
\includegraphics[width=\linewidth]{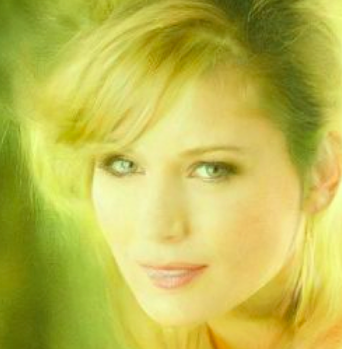} &
\includegraphics[width=\linewidth]{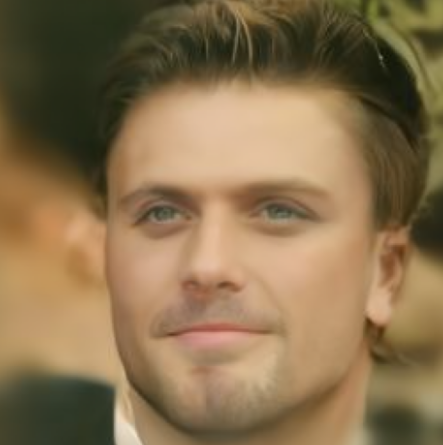} \\[0pt]
{\rotatebox[origin=c]{90}{\footnotesize\textbf{Ours}}} &
\includegraphics[width=\linewidth]{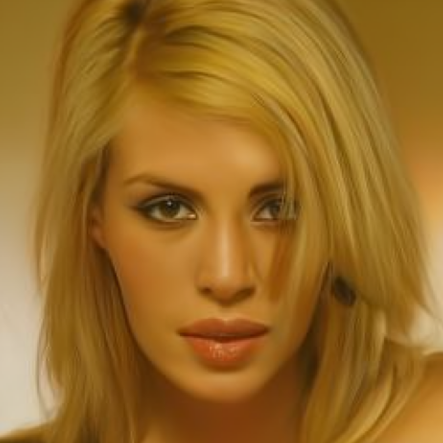} &
\includegraphics[width=\linewidth]{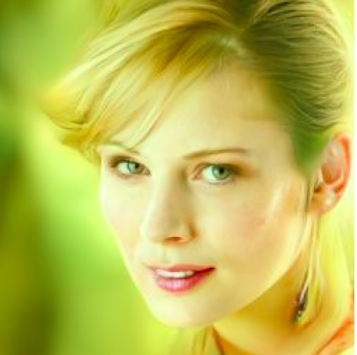} &
\includegraphics[width=\linewidth]{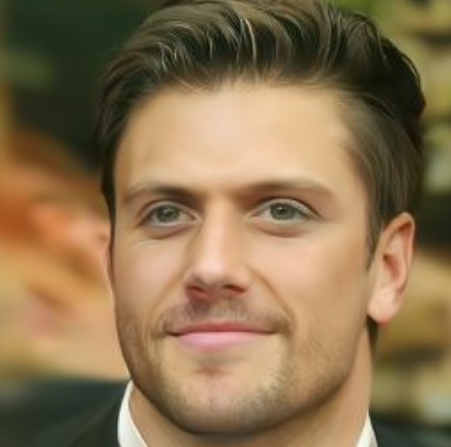} \\
\end{tabular}
\caption{Our method yields sharper details and fewer artifacts than the baseline in all three examples.}
\label{fig:image_grid_2x3}
\vspace{-10pt}
\end{figure}

Despite these advances, fine-grained control over low-level perceptual features such as sharpness, contrast, saturation remain an open problem in unconditional diffusion models. These concepts are well-studied in generative adversarial networks \citep{goodfellow2014generativeadversarialnetworks}. Interpretable directions in the GAN latent space have been shown to correspond to photometric transformations including brightness, color balance, and contrast \citep{jahanian2019steerability,spingarn2021gan,Fruend711804,10.1167/jov.23.10.14}. 

A promising direction comes from the U-Net bottleneck, or \emph{h-space}, which behaves as a semantically dense latent space amenable to linear manipulation \citep{kwon2022semantic,park2023semantic,haas2024hspacepca}. These properties make \emph{h-space} a natural candidate for encoding low-level concept directions.

\begin{figure*}[ht]
 \centering
 \includegraphics[width=0.85\textwidth]{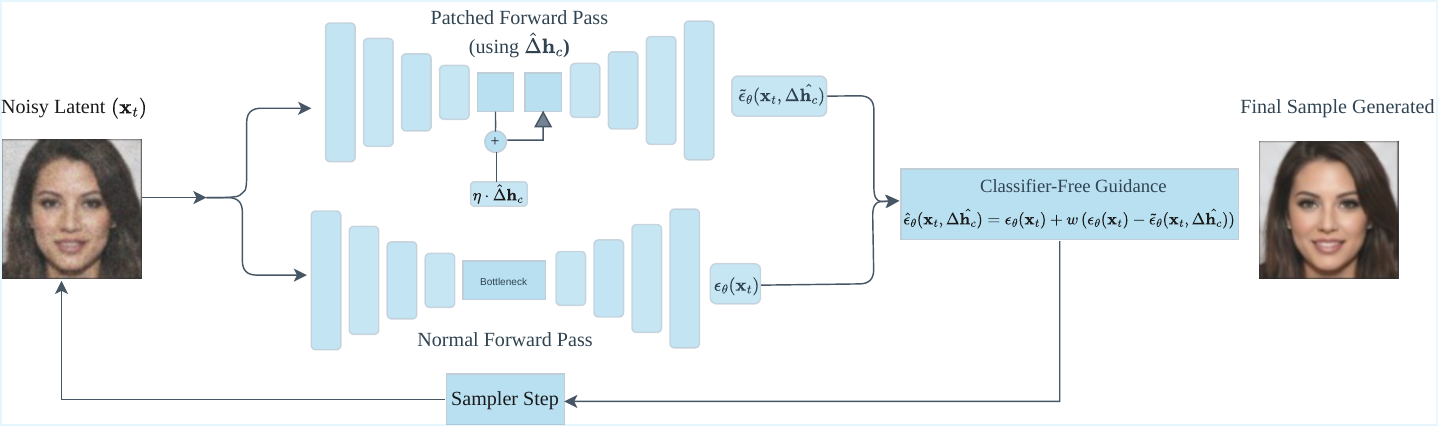}
\caption{Overview of our inference framework. The process begins after the concept vector, shown in \cref{fig:my_diagram}, is computed as the mean pairwise difference of the degraded and clean \emph{h-space} vectors at the extraction timestep $k$. This is followed by (a) Patching where the vector is injected into the bottleneck during inference. Finally, the modified activations are used to compute the (b) Guided Noise Prediction, which is used as the updated prediction of the score function $\nabla_{\mathbf{x}_t}\text{log}\,p(\mathbf{x}_t)$}
 \label{fig:method-overview}
\end{figure*}

We introduce a unified editing paradigm to achieve this. We first extract supervised directions in \emph{h-space} corresponding to low-level degradations (blur, low contrast, grayscale)(\Cref{fig:my_diagram}). After inhibiting this direction in the bottleneck of the U-Net, the structurally degraded noise prediction is used to guide the generative trajectory towards aesthetically enhanced images using classifier-free guidance (Figure~\ref{fig:method-overview}). Since our inference method generalizes \citep{park2023semantic,haas2024hspacepca}, it supports both low-level perceptual and semantic concept editing. We validate through FID comparisons and human evaluation studies (\Cref{tab:combined-results}), consistently outperforming activation patching baselines across low-level editing directions.

We show that classical bottleneck patching fails on this problem due to destructive interference in the decoder of the U-Net. We perform ablations  showing our method's transferability across
datasets. We also experiment with time-dependent guidance
schedules during reverse diffusion to decrease computational cost.

\section{Methodology}
\label{sec:method}
We formalize an inference-time mechanism aimed at guiding the diffusion process away from concepts associated with low-level perceptual degradation. Our approach is: (1) isolating degraded bottleneck feature representations; (2) applying activation patching during inference; and (3) updating the noise predictions through classifier-free guidance.



\subsection{Concept Vector Extraction}
\label{sec:vector_extraction}
\begin{figure}[H]
   \centering
   \captionsetup{skip=0 pt}
   \includesvg[width=0.95\columnwidth]{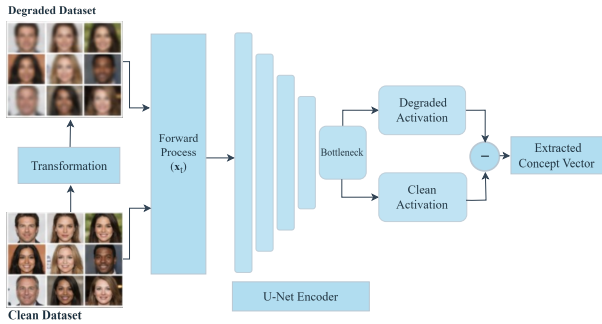}
   \caption{Paired-data concept-vector extraction $\Delta \mathbf{h}_c$.}
   \label{fig:my_diagram}
\end{figure}

Let $\mathcal{T}_c$ denote a transformation\footnote{Mathematical formulas provided in Appendix~\ref{apx:transformation}} causing a degradation $c$ (e.g. blur, grayscale, low-contrast) in a clean image ($\mathbf{x}$). Let $\mathbf{h}_t$ and $\mathbf{h}'_t$ denote the bottleneck activations of $\mathbf{x}$ and $\mathcal{T}_c(\mathbf{x})$ at extraction timestep $t$. The degraded concept vector $\Delta{\mathbf{h}_c}$ is then computed as the average pairwise difference:
\begin{equation}
   \Delta{\mathbf{h}_c} =
   \frac{1}{N}\sum_{i=1}^{N} \left({\mathbf{h'}}_t^{(i)}-\mathbf{h}_t^{(i)}\right)
   \label{eq:vector_extraction}
\end{equation}
This yields a direction $\Delta{\mathbf{h}_c}$ in the \emph{h-space} that points from the clean toward the degraded concept manifold. Fig~\ref{fig:my_diagram} displays the approach.

The choice of timestep $t$ governs the frequency content of the activations used for extraction.\footnote{Further analysis is provided in Appendix~\ref{apx:vector_sep}}

\subsection{Inference Activation Patching}
During the reverse diffusion process, at each timestep $t$, we intercept the bottleneck activation $\mathbf{h}_t$ produced by the
U-Net encoder and apply a directional patch in the direction
of $\hat{\Delta{\mathbf{h}_c}}$:

\begin{equation}
   \tilde{\mathbf{h}}_t = \mathbf{h}_t +
   \eta \cdot {\hat{\Delta{\mathbf{h}_c}}}
   \label{eq:patching}
\end{equation}
where $\hat{\Delta{\mathbf{h}_c}} = \Delta{\mathbf{h}_c} /
\|\Delta{\mathbf{h}_c}\|_2$ is the unit-normalized concept
vector, $\eta \in \mathbb{R}^{+}$ controls the magnitude of the patching. Unlike prior activation patching methods~\citep{kwon2022semantic,gandikota2024conceptsliders}, we patch in a direction \textit{towards} the degradation of the image. We justify this empirically in Section~\ref{sec:analysis}.

\subsection{Negative Classifier-Free Guidance}
We use the noise predicted by the U-Net decoder as a \textit{conditional proxy} for the degraded concept. We then apply classifier-free guidance~\citep{ho2022cfg} to guide away from it by reversing the sign of the guidance scale. At each timestep $t$, the modified noise prediction is:
\begin{equation}
 \begin{split}
 \hat{\epsilon}_\theta(\mathbf{x}_t,\hat{\Delta{\mathbf{h}_c}})
   = \epsilon_\theta(\mathbf{x}_t) + w\,(\epsilon_\theta(\mathbf{x}_t)-\tilde{\epsilon}_\theta(\mathbf{x}_t,\hat{\Delta{\mathbf{h}_c}}))
 \end{split}
 \label{eq:negative_cfg}
\end{equation}
where $\epsilon_\theta(\mathbf{x}_t)$ is the standard unconditional noise prediction, $\tilde{\epsilon}_\theta(\mathbf{x}_t,\hat{\Delta{\mathbf{h}_c}})$ is the noise prediction after intercepting with the patched bottleneck $\tilde{\mathbf{h}}_t$ using \Cref{eq:patching},
and $w \in \mathbb{R}^+$ is the guidance scale.

Figure~\ref{fig:method-overview} summarizes the overall workflow, detailing the corresponding procedure. We also
conduct a thorough inspection of this formulation in
Section~\ref{sec:analysis}.

\section{Experiments}
\label{sec:experiments}

\subsection{Implementation Details and Metrics}
We instantiate our method on a frozen pretrained unconditional DDPM backbone\footnote{HuggingFace checkpoint: \url{https://huggingface.co/google/ddpm-celebahq-256}.}, applying all edits at inference time without updating any model parameters. All experiments use $256\times256$ resolution with 30 DDIM steps on CelebA-HQ. For each degradation concept, vectors are extracted from paired degraded/clean subsets of $N \leq 100$ images. 

\begin{figure*}[!t]
\centering
\setlength{\tabcolsep}{1pt}
\renewcommand{\arraystretch}{1.0}

\resizebox{0.9\textwidth}{!}{%
\begin{tabular}{ccccccc}
  \multicolumn{1}{c}{}
  & \multicolumn{2}{c}{\small\textbf{Sharpness}}
  & \multicolumn{2}{c}{\small\textbf{Saturation}}
  & \multicolumn{2}{c}{\small\textbf{Contrast}} \\[-0.5ex]

    {\scriptsize Baseline}
    & {\scriptsize Standard Patching}
    & {\scriptsize\textbf{Ours}}
    & {\scriptsize Standard Patching}
    & {\scriptsize\textbf{Ours}}
    & {\scriptsize Standard Patching}
    & {\scriptsize\textbf{Ours}} \\[0.5ex]

      \includegraphics[width=2cm]{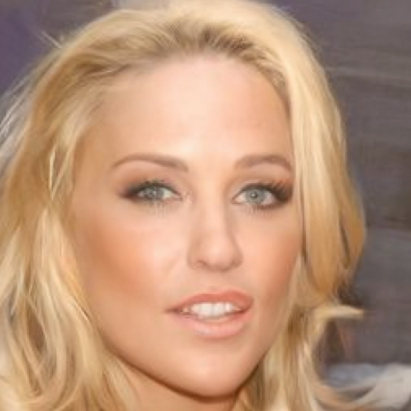}
      & \includegraphics[width=2cm]{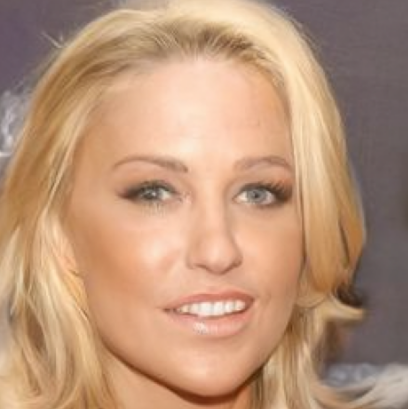}
      & \includegraphics[width=2cm]{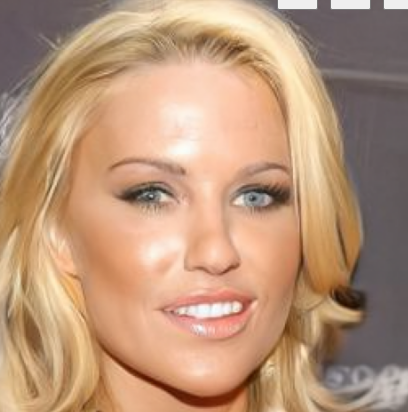}
      & \includegraphics[width=2cm]{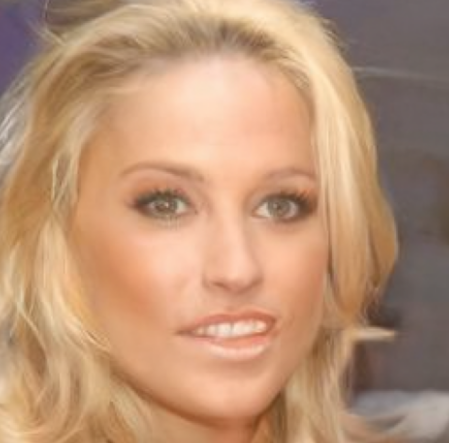}
      & \includegraphics[width=2cm]{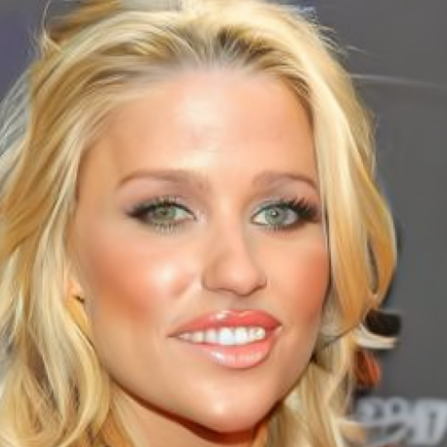}
      & \includegraphics[width=2cm]{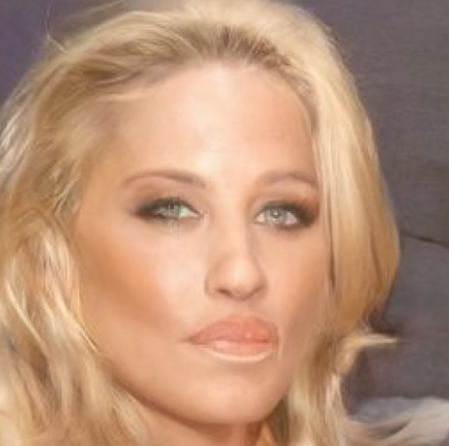}
      & \includegraphics[width=2cm]{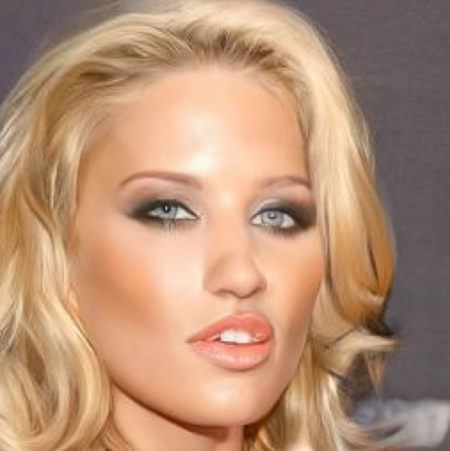} \\[-2pt]

        \includegraphics[width=2cm]{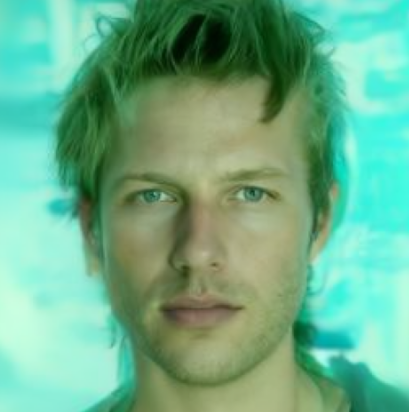}
        & \includegraphics[width=2cm]{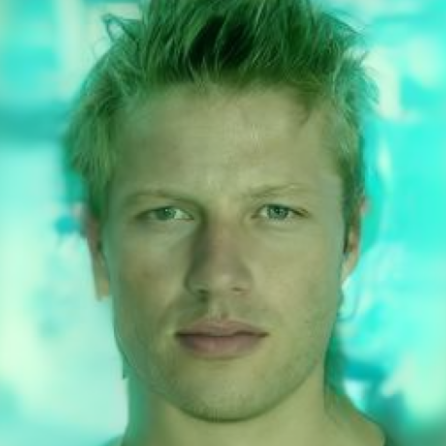}
        & \includegraphics[width=2cm]{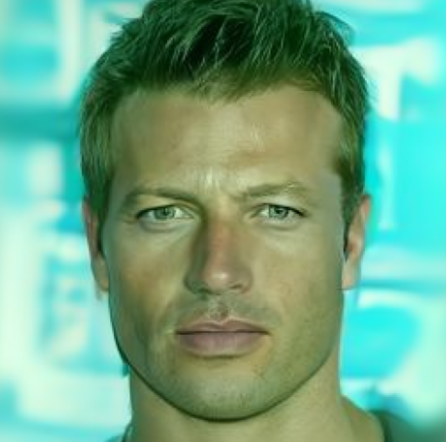}
        & \includegraphics[width=2cm]{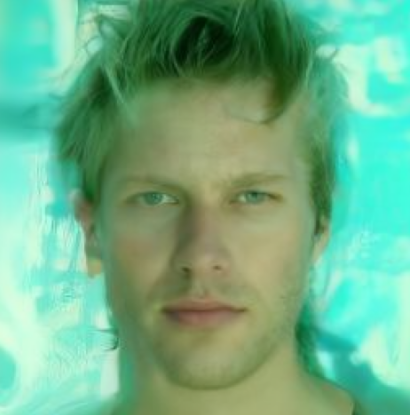}
        & \includegraphics[width=2cm]{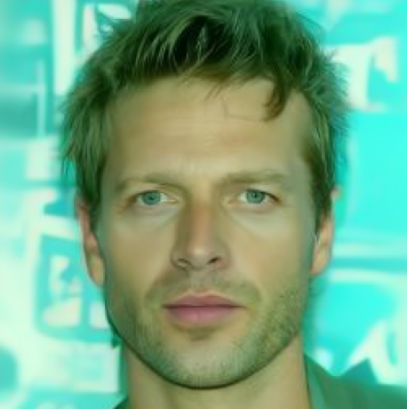}
        & \includegraphics[width=2cm]{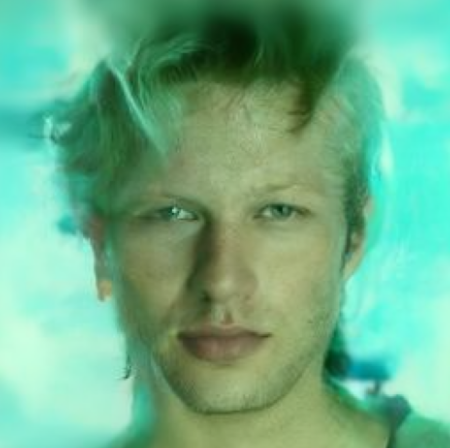}
        & \includegraphics[width=2cm]{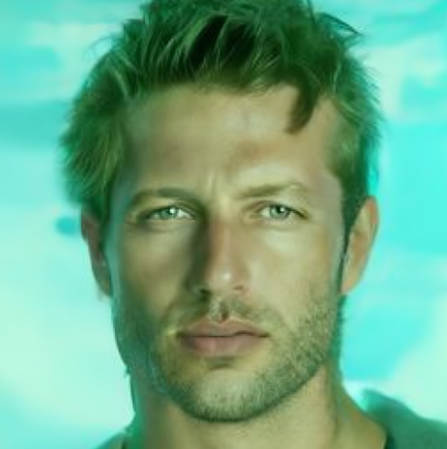} \\[-2pt]

          \includegraphics[width=2cm]{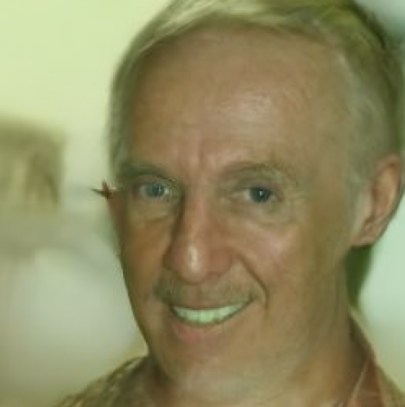}
          & \includegraphics[width=2cm]{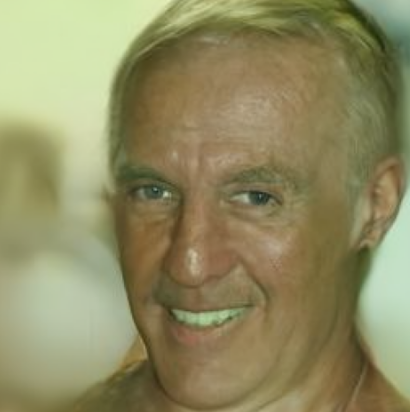}
          & \includegraphics[width=2cm]{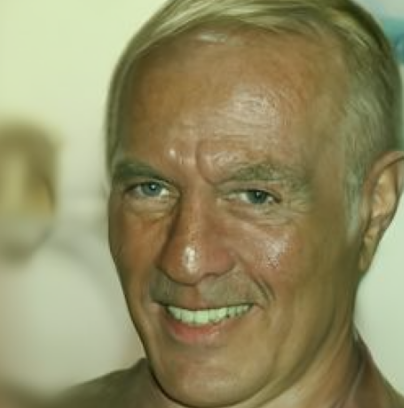}
          & \includegraphics[width=2cm]{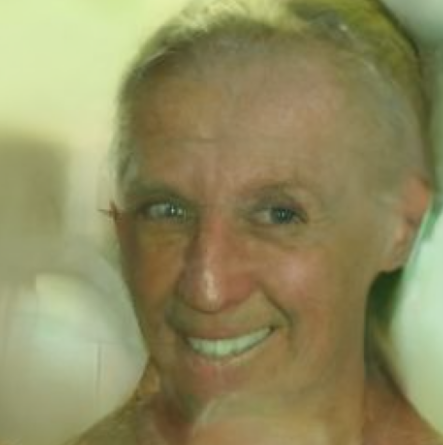}
          & \includegraphics[width=2cm]{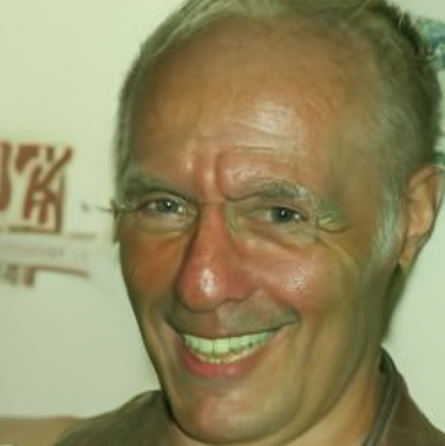}
          & \includegraphics[width=2cm]{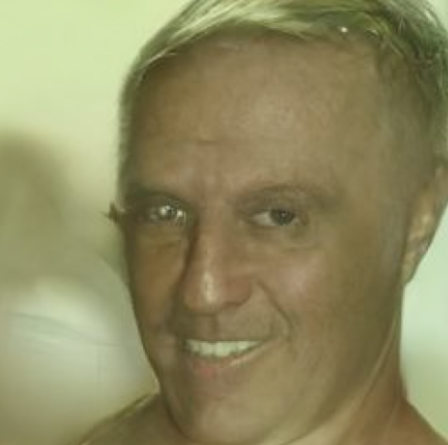}
          & \includegraphics[width=2cm]{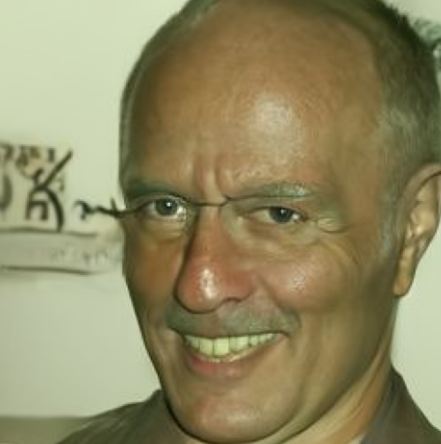} \\
          \end{tabular}}

          \caption{Comparison of Baseline, Standard Patching and our method across sharpness, saturation, and contrast. For Guidance, a CFG-scale
of +1.5 was used for all the samples across all the directions.}
          \label{fig:image_comparison}
          \end{figure*}
We consider three low-level degradations \emph{blur}, \emph{low contrast}, and \emph{grayscale} and report their positive edit directions as \emph{sharpness}, \emph{contrast}, and \emph{saturation}, respectively.

For evaluation, we report Fréchet Inception Distance (FID) and human pairwise A/B comparisons between our method and \citep{haas2024hspacepca} under identical initial noise and sampling settings. 60 human evaluators selected the image better representing the target edit, with the baseline sample as anchor, and a \emph{neither} option available.
\subsection{Results}
\label{sec:results}
Qualitatively, our method produces edits that are stronger and cleaner than standard patching alone (Fig~\ref{fig:image_comparison}). For sharpness, patch-only editing introduces indiscriminate smoothing or structural artifacts, whereas our method yields the intended enhancement while preserving global composition and identity. Similar trends hold for contrast and saturation, where patch-only editing produces washed-out highlights or uneven desaturation compared to our more uniform transformations.

Quantitatively, Table~\ref{tab:combined-results} confirms these observations. Across all concepts, our method achieves lower FID and higher direction-specific metrics: Laplacian variance for sharpness, mean S-channel for saturation, and RMS contrast for contrast. Human evaluations corroborate this: annotators prefer our method in \textbf{76.0\%} of pairwise comparisons for sharpness, with consistent advantages across remaining concepts. Together, these results demonstrate that combining bottleneck patching with noise-space extrapolation outperforms either component in isolation.\footnote{Baseline FID is comparatively high due to limited compute, which constrained the number of generated samples to 10k per concept direction and inference steps to 30 (DDIM) used for evaluation.}

\begin{table}[ht]
    \centering
    \caption{Percentage change relative to baseline FID, where negative values 
    indicate improvement. Direction-specific metrics report absolute values, higher is better.}
    \label{tab:combined-results}
    \small
    {\setlength{\tabcolsep}{4pt}%
    \renewcommand{\arraystretch}{1.4}%
    \resizebox{0.96\columnwidth}{!}{%
    \begin{tabular}{@{}llccc@{}}
    \toprule
    Direction & Metric & Baseline & Standard Patching & Ours \\
    \midrule
    Sharpness & FID (\%$\Delta$) & 25.43 & +3.54\% & \textbf{-6.07\%} \\
              & Laplacian variance & 143.77 & 212.72 & \textbf{386.45} \\
    \midrule
    Saturation & FID (\%$\Delta$) & 25.43 & +7.03\% & \textbf{-7.76\%} \\
               & Mean S-channel & 0.43 & 0.44 & \textbf{0.47} \\
    \midrule
    Contrast & FID (\%$\Delta$) & 25.43 & +4.85\% & \textbf{-13.90\%} \\
             & RMS contrast & 0.18 & 0.17 & \textbf{0.21} \\
    \bottomrule
    \end{tabular}%
    }}
\end{table}

Further ablations validating generalization, semantic concept directions, 
and our partial guidance variant are provided in \cref{apx:ablations}.
\subsection{Empirical Analysis of Patching and Guidance}
\label{sec:analysis}

Empirically validating Sec~\ref{sec:method}, we demonstrate that the \emph{h-space} linearity assumed in prior work \citep{kwon2022semantic, haas2024hspacepca, park2023semantic} fails for low-level perceptual degradation. Unlike high-level features, negatively patching our concept vector into the U-Net bottleneck causes destructive interference and collapses image quality. However, our method remains stable, concentrating the sampling distribution toward viable, upgraded outputs.

\begin{figure}[t]
    \centering
    \makebox[\columnwidth][c]{\includegraphics[width=0.95\columnwidth]{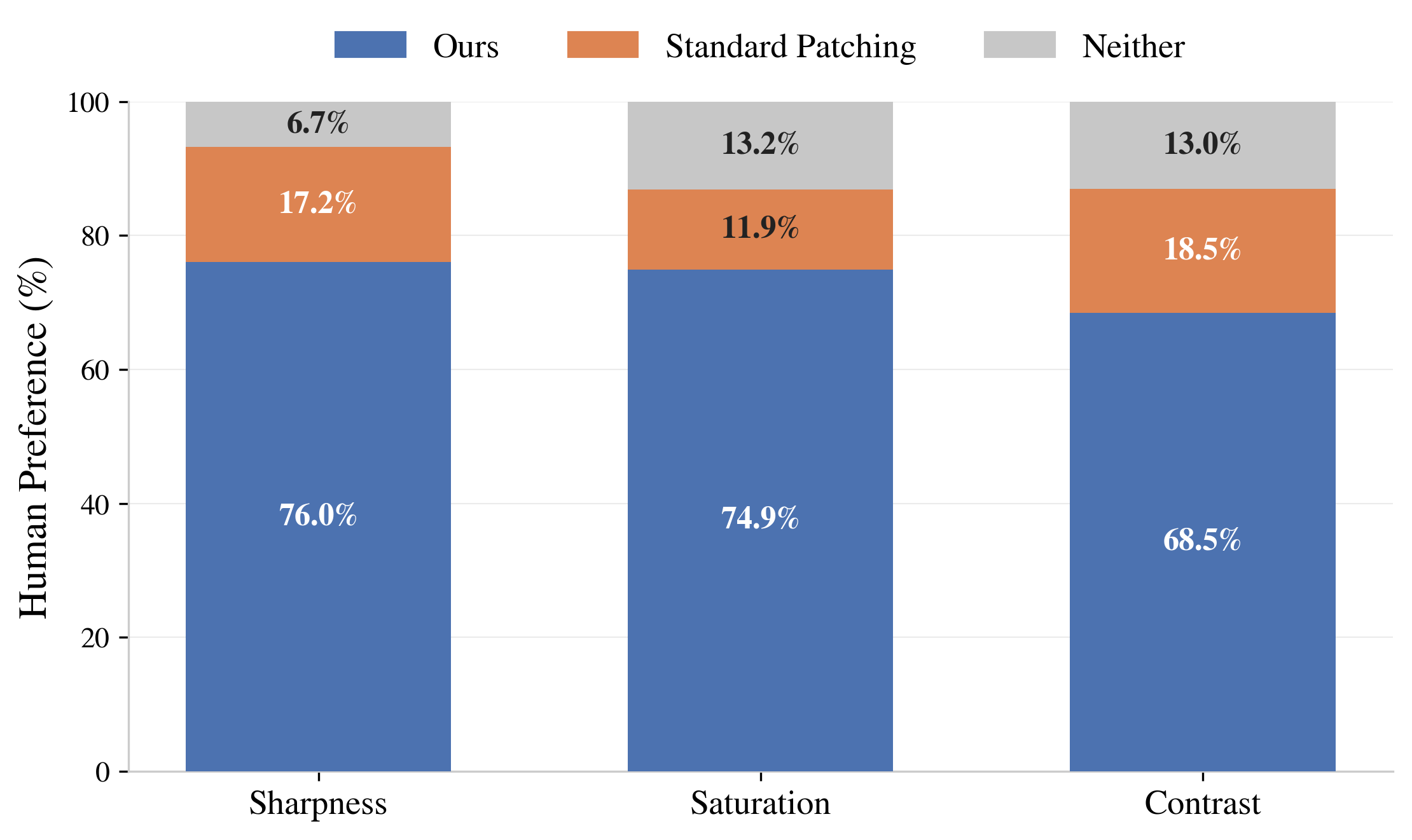}}
    \caption{Human preference rates across sharpness, saturation, and
    contrast. Our method is preferred in all three settings.}
    \label{fig:human-preference}
\end{figure}

\textbf{Non-linear disturbance in the decoder}: We express the noise prediction $\hat{\epsilon}_\theta$ (\ref{eq:negative_cfg}) in terms of the U-Net decoder $\mathcal{D}_\theta$, which is written as a function of the bottleneck activation $\mathbf{h}_t$, skip-connections $\{s_i\}_{i=1}^{L}$, timestep embedding $(\tau)$ and concept vector $\Delta\hat{\mathbf{h}}_c$. 

The skip-connections $\{s_i\}$ and timestep 
embedding $\tau$ are suppressed from the notation for clarity.
\begin{equation*}
    \hat{\epsilon}_\theta\!\left(\mathbf{x}_t, \Delta\hat{\mathbf{h}}_c\right)
    = \mathcal{D}_\theta(\mathbf{h}_t)
    + w\cdot\Bigl[
        \mathcal{D}_\theta(\mathbf{h}_t)
        - \mathcal{D}_\theta(\mathbf{h}_t + \eta\Delta\hat{\mathbf{h}}_c)
    \Bigr]
    \label{eq:cfg_decoder}
\end{equation*}

Since $\mathcal{D}_\theta$ is composed of differentiable 
operations,\footnote{Architecture details: 
\href{https://huggingface.co/google/ddpm-celebahq-256}
{\texttt{google/ddpm-celebahq-256}}.} 
we expand $\mathcal{D}_\theta(\mathbf{h}_t + \eta\,\Delta\hat{\mathbf{h}}_c)$ 
in a Taylor series about $\mathbf{h}_t$:
\begin{equation}
  \mathcal{D}_\theta\!\left(\mathbf{h}_t + \eta\,\Delta\hat{\mathbf{h}}_c\right)
    = \mathcal{D}_\theta(\mathbf{h}_t)
    + \eta\,\mathbf{J}_{\!\mathcal{D}}(\mathbf{h}_t)\,\Delta\hat{\mathbf{h}}_c
    + {\mathbf{R}_{\geq 2}}
    \label{eq:taylor}
\end{equation}
where $\mathbf{J}_{\!\mathcal{D}}(\mathbf{h}_t)\,\Delta\hat{\mathbf{h}}_c$ is the Jacobian-vector product, and $\mathbf{R}_{\geq 2}$ collects all higher-order terms.

We obtain the decomposed update equation as:
\begin{equation}
    \hat{\epsilon}_\theta\!\left(\mathbf{x}_t, \Delta\hat{\mathbf{h}}_c\right)= \mathcal{D}_\theta(\mathbf{h}_t)
    - w\cdot\eta\;
      \mathbf{J}_{\!\mathcal{D}}(\mathbf{h}_t)\Delta\hat{\mathbf{h}}_c -w\cdot \mathbf{R}_{\geq 2}
    \label{eq:decomposed}
\end{equation}

To highlight the non-linearity present in the decoder, which acts as the cause of failure in negative patching, we track the \emph{relative residual ratio}:
\begin{equation}
    \rho \;=\; \frac{\|w\cdot \mathbf{R}_{\geq 2}\|_2}{\|\hat{\epsilon}_\theta\!\left(\mathbf{x}_t, \Delta\hat{\mathbf{h}}_c\right) - \mathcal{D}_\theta(\mathbf{h}_t)\|_2}
    \label{eq:ratio}
\end{equation}
Relative residual ratio $(\rho)$ quantifies how well the linear term $(\hat\epsilon_\theta^{1})$ aligns with the nonlinear components in~(\ref{eq:decomposed}).\footnote{A complete derivation is provided in Appendix~\ref{apx:residual}.}
\begin{figure}[ht]
  \centering
  \includegraphics[width=0.9\columnwidth]{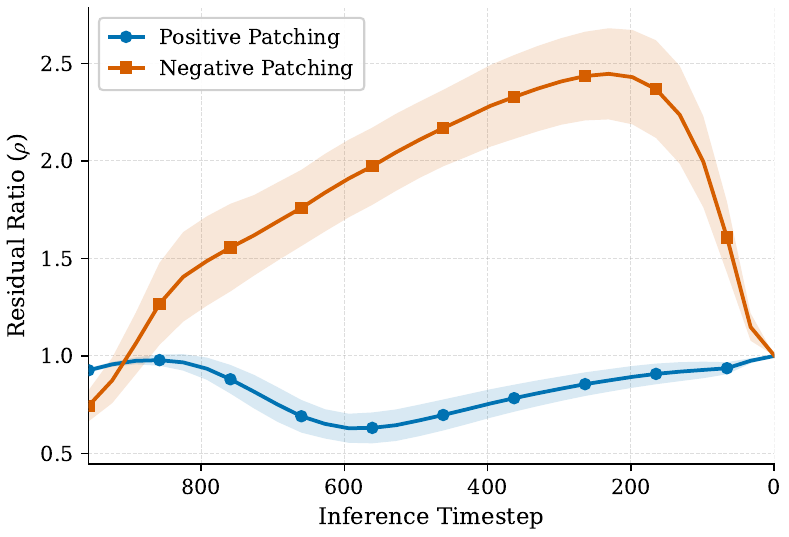}
  \caption{Relative residual ratio $\rho$ across denoising timesteps for the blur direction.}
\end{figure}

When patching positively ($\eta>0$), $\rho < 1$ throughout the trajectory, confirming that higher-order terms remain aligned with the guidance direction, preserving image quality. Conversely, negative patching ($\eta < 0$) causes $\rho$ to overshoot unity mid-trajectory, inducing destructive interference with the first-order term, where the decoder's non-linear response opposes the intended guidance direction and degrades output quality.

Additionally, we observed that direct patching in the negative direction blurs the decoder's attention maps. This phenomenon is analyzed in detail in the \cref{apx:entropy}.

\section{Future Work}
Several directions remain open. More principled approaches to reducing computational cost, such as adaptive guidance scheduling driven by real-time signals like attention entropy, could make the method practical at larger scale. Extending the framework to flow-based generative models and text-conditioned diffusion models would significantly broaden its applicability. Replacing manual hyperparameter selection with self-tuning mechanisms that infer appropriate patching and guidance strengths directly from image content would improve usability. Finally, a sophisticated, theoretical study on failure of traditional patching with low-level transformations constitutes an interesting direction of work.
\section{Conclusion}
We presented a training-free, inference-time framework for inducing low-level perceptual transformations in unconditional diffusion models, successfully producing improvements without any model retraining. Our pipeline combined two separately effective techniques in the field of diffusion models: \emph{h-space} manipulation and classifier-free guidance. We further provided a mechanistic analysis of why prior methods fail for low-level edits, tracing the breakdown to unstable non-linear residual in the decoder of the U-Net. Ablations across datasets and concept directions confirm the generality of our approach.
\newpage


{
    \small
    \bibliographystyle{ieeenat_fullname}
    \bibliography{main}
}

\clearpage
\appendix
\onecolumn
\begin{center}
   \textbf{\Large Appendix} \\
   \vspace{1em}
   \vspace{2em}
\end{center}

\section{Transformations}
\label{apx:transformation}

For each contrastive pair, the degraded image is obtained by applying the transformation to the clean RGB image prior to resizing and normalization. The three transformations are defined as follows.

\paragraph{Blur.}
The blurred image $\tilde{x}$ is obtained by convolving the clean image $x$ with a 2-D Gaussian kernel $K$:

\begin{equation}
    \tilde{x}[i,j] = \sum_{m=-r}^{r} \sum_{n=-r}^{r} K[m,n] \cdot x[i+m,\, j+n]
\end{equation}

\noindent where the kernel is constructed as the outer product of a 1-D Gaussian:

\begin{equation}
    K = \mathbf{k}\mathbf{k}^{\top}, \qquad k_i = \frac{1}{Z}\exp\!\left(-\frac{i^2}{2\sigma^2}\right)
\end{equation}

\noindent with kernel size $21 \times 21$, standard deviation $\sigma = 3.0$, half-width $r = 10$, and $Z$ being the normalisation constant. Border pixels are handled via reflect padding.

\paragraph{Grayscale.}
The image is converted to luminance using the ITU-R BT.601 luma coefficients and then replicated across all three channels:

\begin{equation}
    \tilde{x} = 0.299\, R + 0.587\, G + 0.114\, B
\end{equation}

\noindent yielding a three-channel tensor $(\tilde{x},\, \tilde{x},\, \tilde{x})$ with no chromatic information.

\paragraph{Low Contrast.}
Contrast is reduced by linearly interpolating each pixel toward the mean luminance $\mu$ of the image:

\begin{equation}
    \tilde{x} = \mu + \alpha \cdot (x - \mu)
\end{equation}

\noindent where $\alpha = 0.6$ is the contrast factor and $\mu$ denotes the per-image mean luminance. Since $\alpha < 1$, the pixel range is compressed toward the mean, attenuating contrast.
\section{Empirical Analysis of Standard Patching and Guidance}

\subsection{Residual-Ratio Derivation}
\label{apx:residual}
Let $\hat{\boldsymbol{\epsilon}}_1 = -w \cdot \eta\,\mathbf{J}_{\!\mathcal{D}}(\mathbf{h}_t)\Delta\hat{\mathbf{h}}_c$
denote the first-order guidance term and $\widehat{\mathbf{R}}_{\geq 2} = -w\cdot\mathbf{R}_{\geq 2}$
denote the scaled higher-order residual from Eq.~\eqref{eq:decomposed}, so that:
\begin{equation}
    \hat{\epsilon}_\theta\!\left(\mathbf{x}_t, \Delta\hat{\mathbf{h}}_c\right) - \mathcal{D}_\theta(\mathbf{h}_t)
    = \hat{\boldsymbol{\epsilon}}_1 + \widehat{\mathbf{R}}_{\geq 2}
    \label{eq:delta}
\end{equation}
The relative residual ratio is then:
\begin{equation}
    \rho = \frac{\|\widehat{\mathbf{R}}_{\geq 2}\|_2}
               {\|\hat{\epsilon}_\theta\!\left(\mathbf{x}_t,\,\Delta\hat{\mathbf{h}}_c\right)
                 - \mathcal{D}_\theta(\mathbf{h}_t)\|_2}
           = \frac{\|\widehat{\mathbf{R}}_{\geq 2}\|_2}
                  {\|\hat{\boldsymbol{\epsilon}}_1 + \widehat{\mathbf{R}}_{\geq 2}\|_2}
    \label{eq:rho}
\end{equation}
Whenever $\rho > 1$, the numerator exceeds the denominator:
\begin{equation}
    \|\widehat{\mathbf{R}}_{\geq 2}\|_2 > \|\hat{\boldsymbol{\epsilon}}_1 + \widehat{\mathbf{R}}_{\geq 2}\|_2
    \label{eq:rho_gt1}
\end{equation}
For this inequality to hold, the two terms $\hat{\boldsymbol{\epsilon}}_1$ and
$\widehat{\mathbf{R}}_{\geq 2}$ must be \emph{mutually opposing} i.e.\ they must have a negative inner product and $\widehat{\mathbf{R}}_{\geq 2}$
must be the larger of the two in norm.
To see this formally, square both sides of Eq.~\eqref{eq:rho_gt1} and
expand the right-hand side:
\begin{align}
    \|\widehat{\mathbf{R}}_{\geq 2}\|^2
    &> \|\hat{\boldsymbol{\epsilon}}_1 + \widehat{\mathbf{R}}_{\geq 2}\|^2 \notag\\
    &= \|\hat{\boldsymbol{\epsilon}}_1\|^2
     + \|\widehat{\mathbf{R}}_{\geq 2}\|^2
     + 2\langle\hat{\boldsymbol{\epsilon}}_1,\,\widehat{\mathbf{R}}_{\geq 2}\rangle
\end{align}
Cancelling $\|\widehat{\mathbf{R}}_{\geq 2}\|^2$ from both sides gives:
\begin{equation}
    0 > \|\hat{\boldsymbol{\epsilon}}_1\|^2
      + 2\langle\hat{\boldsymbol{\epsilon}}_1,\,\widehat{\mathbf{R}}_{\geq 2}\rangle
\end{equation}
and therefore:
\begin{equation}
    \langle\hat{\boldsymbol{\epsilon}}_1,\,\widehat{\mathbf{R}}_{\geq 2}\rangle
    < -\tfrac{1}{2}\|\hat{\boldsymbol{\epsilon}}_1\|^2 < 0
    \label{eq:inner_product}
\end{equation}
Equation~\eqref{eq:inner_product} establishes two simultaneous conditions.
First, the inner product is strictly negative, confirming that
$\widehat{\mathbf{R}}_{\geq 2}$ is directionally antagonistic to
$\hat{\boldsymbol{\epsilon}}_1$: the nonlinear decoder response actively
opposes the intended first-order guidance direction.
Second, the magnitude of this opposition is bounded below by
$\tfrac{1}{2}\|\hat{\boldsymbol{\epsilon}}_1\|^2$, which grows with the
guidance scale $w$ and patching magnitude $\eta$ the stronger the
intended guidance, the more aggressively the nonlinear residual resists it.
Therefore, whenever $\rho > 1$, the decoder's higher-order response is
not merely large in magnitude; it is geometrically antagonistic to the
linear guidance signal, actively inverting the intended update direction
in noise space and destabilising the denoising trajectory.

\subsection{Additional Diagnostic Plots}
\label{apx:additional_plots}

\begin{figure}[H]
    \centering
    \begin{subfigure}{0.48\textwidth}
        \centering
        \includegraphics[width=\linewidth]{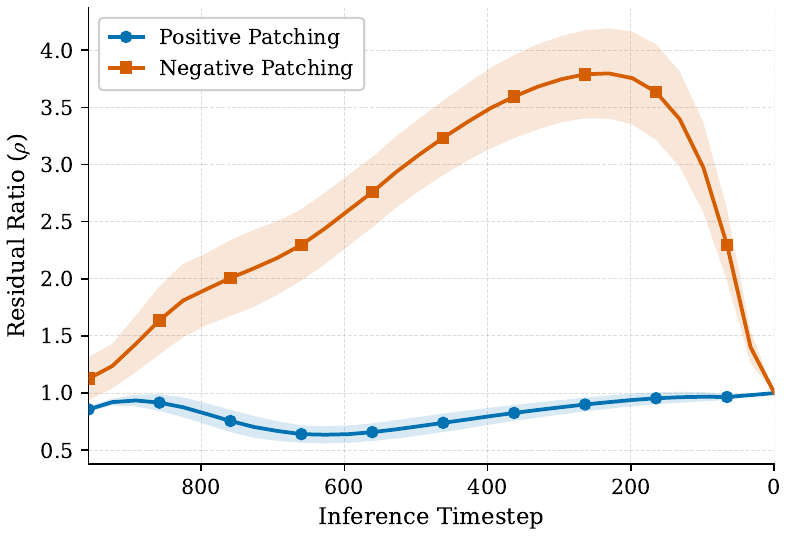}
        \caption{Contrast direction.}
    \end{subfigure}\hfill
    \begin{subfigure}{0.48\textwidth}
        \centering
        \includegraphics[width=\linewidth]{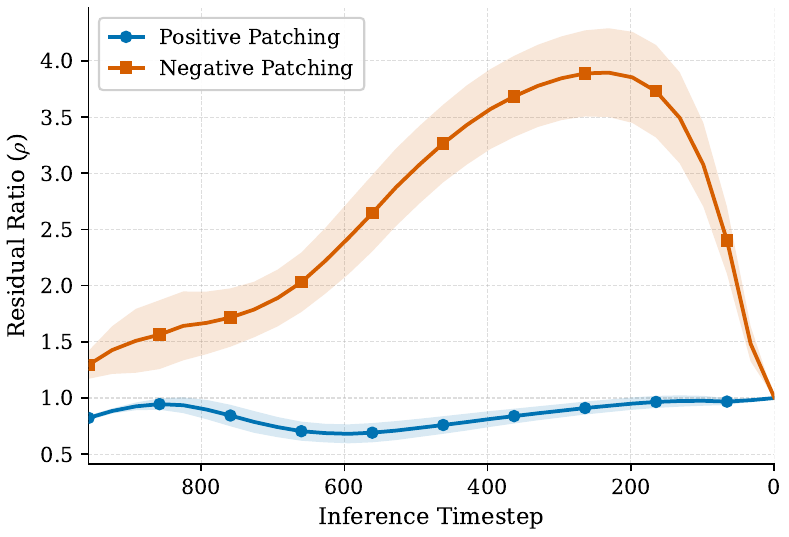}
        \caption{Color direction.}
    \end{subfigure}
    \caption{Residual ratios for the additional learned
    directions. As in the blur case, positive patching yields lower
    entropy than negative patching across most of the denoising
    trajectory, indicating more coherent skip bottleneck alignment.}
    \label{fig:other_residual_dir}
\end{figure}

\subsection{Attention Entropy Analysis}
\label{apx:entropy}
After bottleneck patching, denoted by
$\tilde{\mathbf{h}}_t = \mathbf{h}_t + \eta\,\hat{\Delta \mathbf{h}}_c$,
the U-Net decoder upsamples and concatenates this representation with the
corresponding encoder skip connections $\{s_i\}$. Since these skip
connections are computed from the unpatched encoder, they preserve the
features of the original mid-trajectory latent $\mathbf{x}_t$. If the
patched bottleneck is poorly aligned with these skip features, the
decoder must reconcile inconsistent information.

To quantify the resulting structural mismatch, we measure the attention
entropy across denoising time inside the decoder block that contains
self-attention in our architecture. Lower entropy indicates a more
focused and coherent attention pattern, while higher entropy reflects
greater uncertainty.

\begin{figure}[ht]
    \centering
    \begin{subfigure}{0.32\textwidth}
        \centering
        \includegraphics[width=\columnwidth]{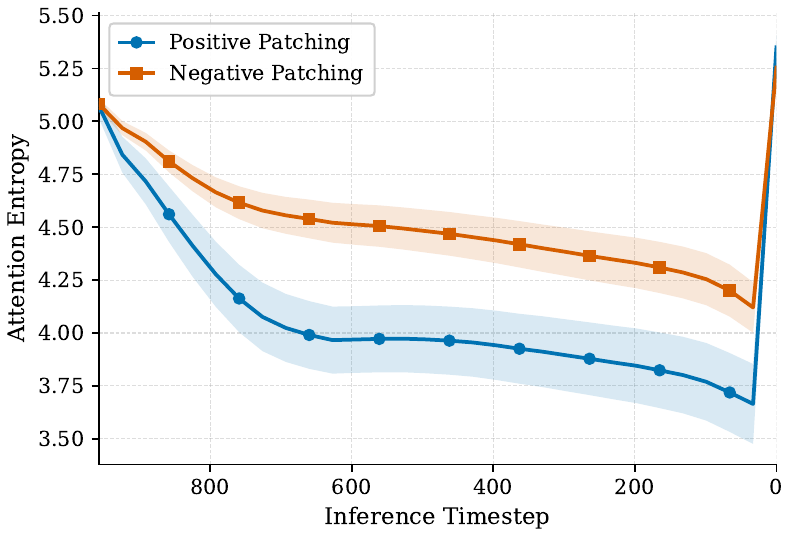}
        \caption{Sharpness direction.}
    \end{subfigure}\hfill
    \begin{subfigure}{0.32\textwidth}
        \centering
        \includegraphics[width=\linewidth]{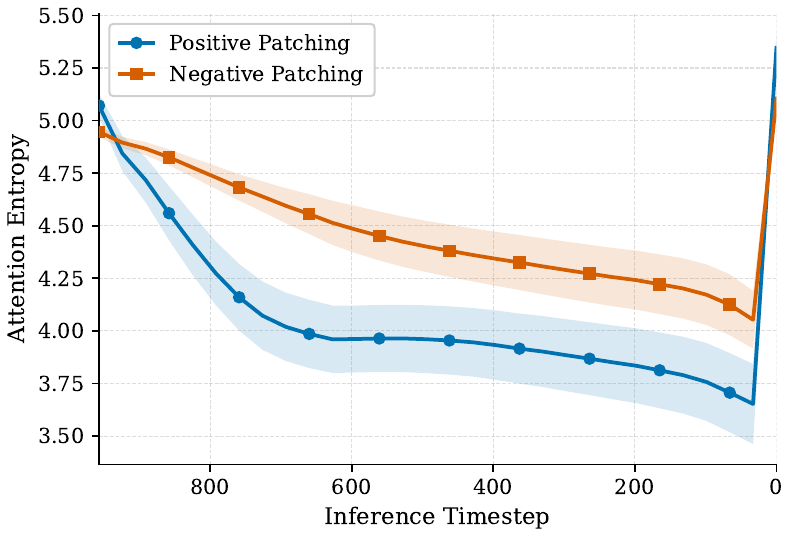}
        \caption{Contrast direction.}
    \end{subfigure}\hfill
    \begin{subfigure}{0.32\textwidth}
        \centering
        \includegraphics[width=\linewidth]{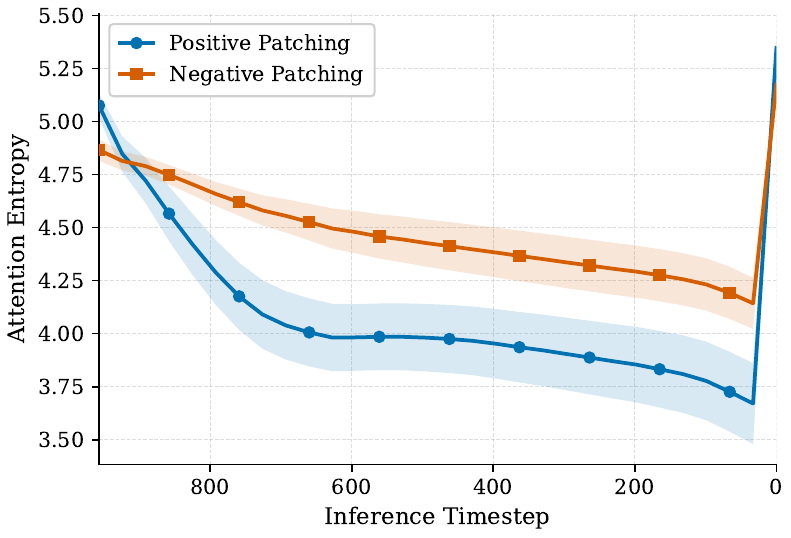}
        \caption{Saturation direction.}
    \end{subfigure}
    \caption{Attention entropy across denoising timesteps for positive and negative bottleneck patching under the sharpness, contrast, and color directions. Across all three directions, negative patching produces higher entropy, indicating less focused attention and poorer skip--bottleneck alignment.}
    \label{entropy_comp}
\end{figure}

\textbf{Positive patching ($\eta > 0$).}
Patching in the degradation direction yields consistently lower
attention entropy throughout the denoising trajectory, indicating better
alignment between bottleneck and skip features.

\textbf{Negative patching ($\eta < 0$).}
\cref{entropy_comp} shows persistently higher attention entropy.
Injecting the opposite direction creates a mismatch with the skip
connections and increases decoder uncertainty.

\section{Concept Separability Across Timesteps}
\label{apx:vector_sep}

To identify the denoising timestep at which a given concept is most
linearly encoded in \emph{h-space}, we measure the linear separability
between clean and degraded bottleneck activations using Linear Discriminant
Analysis (LDA)~\citep{fisher1936use}. Concretely, we compute the Fisher
criterion $\mathcal{F}(t)$ at each timestep $t$:

\begin{equation}
    \mathcal{F}(t) = \frac{(w^{*\top}(\mu^+ - \mu^-))^2}
                         {w^{*\top} S_W\, w^*}
    \label{eq:lda_score}
\end{equation}
\begin{figure}[ht]
    \centering
    \begin{subfigure}{0.32\textwidth}
        \centering
        \includegraphics[width=\linewidth]{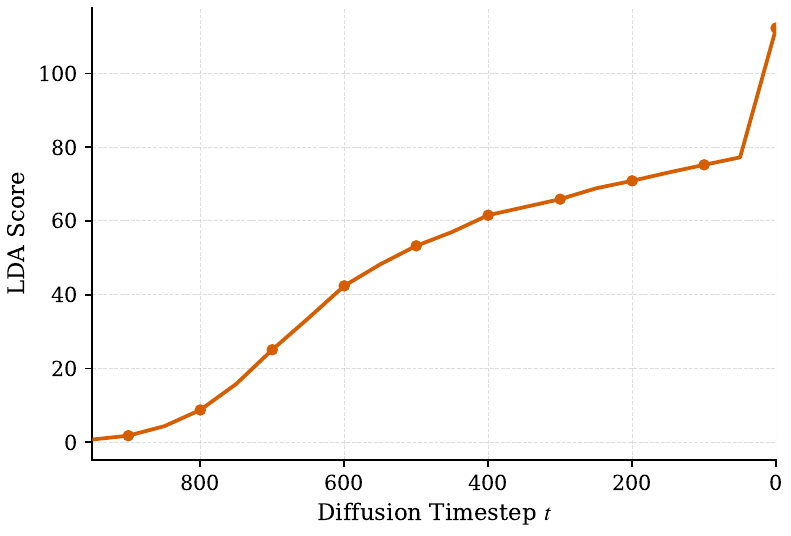}
        \caption{Sharpness direction.}
    \end{subfigure}\hfill
    \begin{subfigure}{0.32\textwidth}
        \centering
        \includegraphics[width=\linewidth]{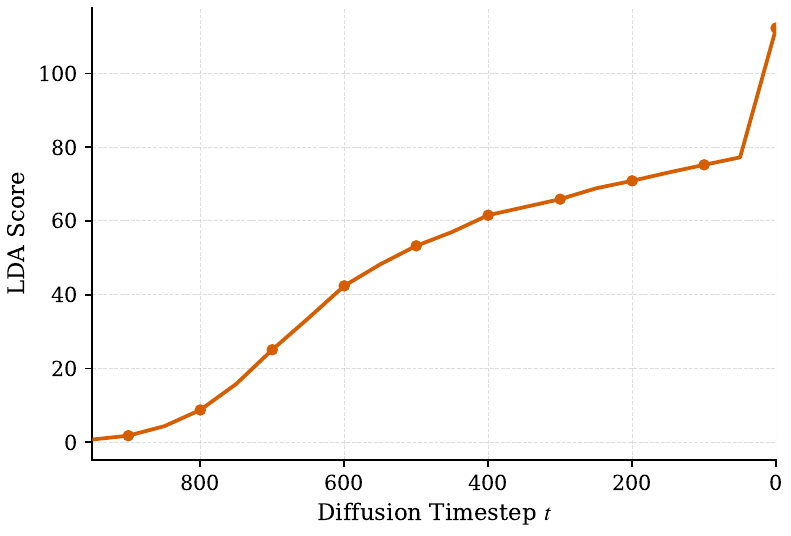}
        \caption{Contrast direction.}
    \end{subfigure}\hfill
    \begin{subfigure}{0.32\textwidth}
        \centering
        \includegraphics[width=\linewidth]{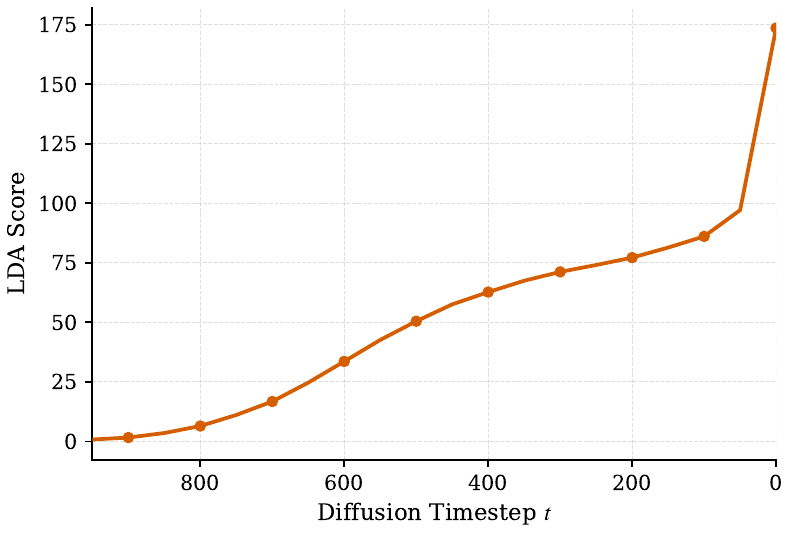}
        \caption{Saturation direction.}
    \end{subfigure}
    \caption{LDA scores across extraction timesteps for the sharpness,
    contrast, and saturation directions.}
    \label{fig:sharpness-lda}
\end{figure}
\noindent where $w^* = S_W^{-1}(\mu^+ - \mu^-)$ is the optimal
discriminant direction, $\mu^+$, $\mu^-$ are the class means, and $S_W$
is the within-class scatter. A higher $\mathcal{F}(t)$ indicates that the
two classes are more discriminable along a linear direction in \emph{h-space}
at timestep $t$. We use $\mathcal{F}(t)$ to identify $t^*$, the timestep
at which concept structure is most cleanly isolated, motivating both our
choice of extraction timestep (\cref{sec:vector_extraction}) and the
selective guidance window. (\cref{selective_guidance}).

\section{Ablations}
\label{apx:ablations}

We validate three core aspects of our method: generalization beyond the training 
domain, applicability to semantic concept directions, and a compute-efficient 
variant that reduces the overhead of continuous guidance.
Unless stated otherwise, all ablations use the same hyperparameters as the 
main experiments.

\subsection{Robustness to Dataset Variation}
\begin{figure}[H]
\centering
\setlength{\tabcolsep}{0pt}
\renewcommand{\arraystretch}{1.1}
\captionsetup{skip=2pt}
\resizebox{0.9\columnwidth}{!}{%
\begin{tabular}{@{}>{\centering\arraybackslash}m{0.075\columnwidth}@{\hspace{1.5pt}}>{\centering\arraybackslash}m{0.298\columnwidth}@{\hspace{1.5pt}}>{\centering\arraybackslash}m{0.298\columnwidth}@{\hspace{1.5pt}}>{\centering\arraybackslash}m{0.298\columnwidth}@{}}

& {\footnotesize{Baseline}} & {\footnotesize{Standard Patching}} & {\footnotesize\textbf{Ours}} \\

& \includegraphics[width=\linewidth]{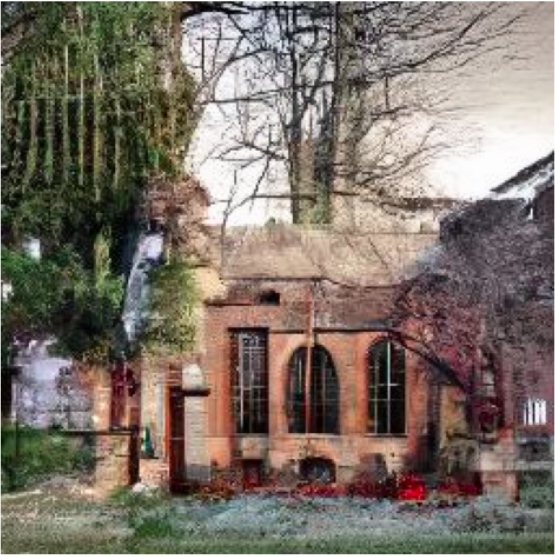} 
& \includegraphics[width=\linewidth]{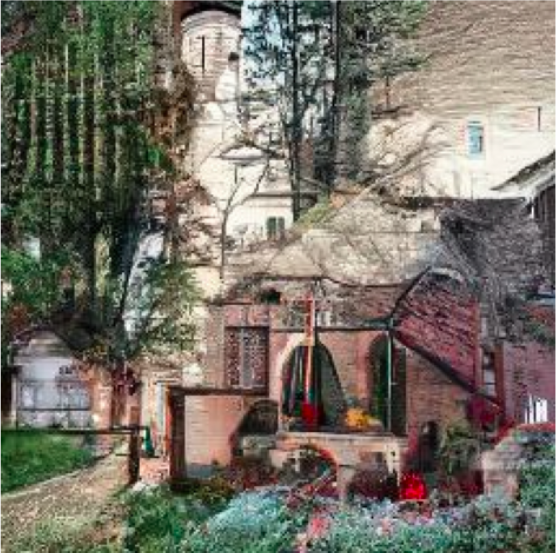} 
& \includegraphics[width=\linewidth]{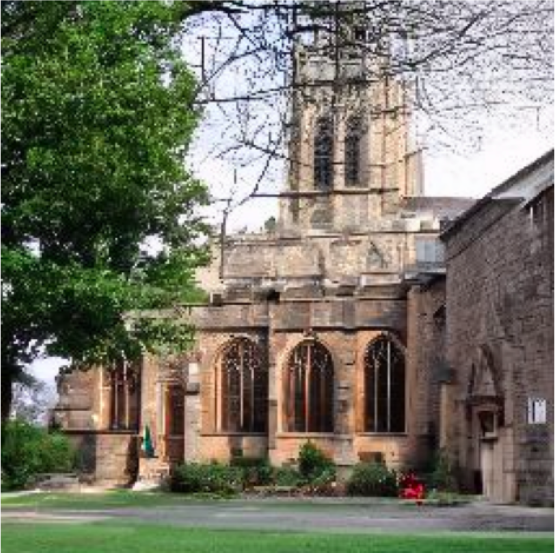} \\[2pt]

& \includegraphics[width=\linewidth]{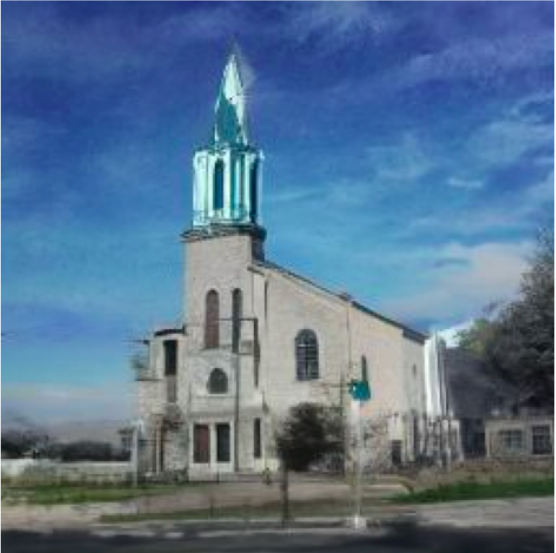} 
& \includegraphics[width=\linewidth]{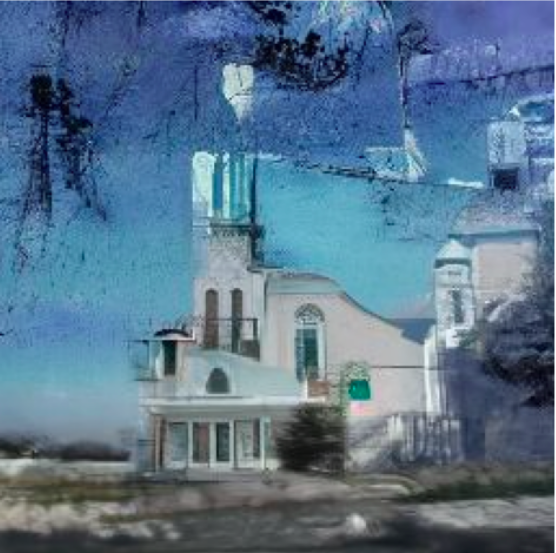} 
& \includegraphics[width=\linewidth]{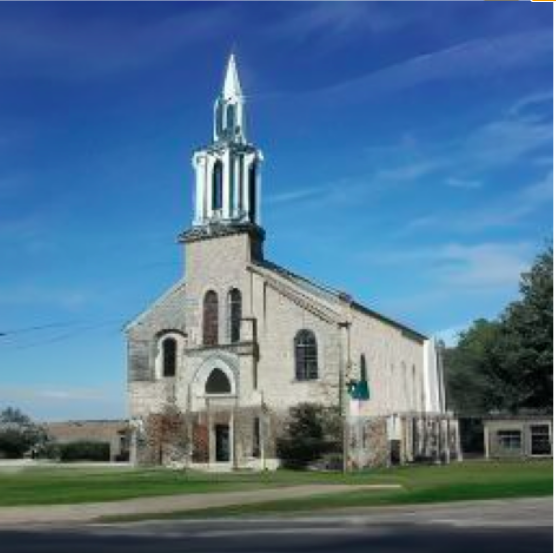} \\

\end{tabular}%
}
\caption{Qualitative comparison between the Baseline, Standard Patching and our Method on three representative examples on LSUN church with Blur direction.}
\label{fig:image_grid_2x3_lsun}
\vspace{-8pt}
\end{figure}

Our main experiments use Celeba-HQ, a face-specific dataset.
To verify that our method is not tailored to face imagery, we test on LSUN Church using the same methodology (Section~\ref{sec:method}).
Figure~\ref{fig:image_grid_2x3_lsun} shows qualitative comparisons between baseline, Standard Patching, and our method with guidance towards direction of degradation on church images.

Our method consistently produces sharper, more detailed images than the baseline,
while standard patching with negative direction at matched $|\eta|$ introduces the same structural
artifacts observed on CelebA-HQ confirming that the failure mode of
negative patching and the benefit of  CFG guidance are not
artifacts of face-specific \emph{h-space} geometry but reflect a general property
of the decoder's directional sensitivity.

\subsection{Semantic Concept Directions.}
\label{apx:semantic_dir}
While our primary contribution targets perceptual-quality attributes,
our framework is agnostic to the choice of concept direction $\Delta\hat{\mathbf{h}}_c$.
\paragraph{Semantic Concept Directions.}
While our primary contribution targets perceptual-quality attributes, our framework is agnostic to the choice of concept direction $\Delta\hat{\mathbf{h}}_c$.
To validate this, we extract semantic directions from CelebA-HQ attribute
labels~\citep{liu2015deep} using a stratified difference-of-means procedure,%
\footnote{We maintain equal proportions of positive and negative examples in each subset.}
focused on the \textsc{Smiling} and \textsc{Male} attributes.
Since synthetic degradation cannot be applied to semantic concepts, we follow the
paired extraction protocol of ~\citep{haas2024hspacepca}, computing
$\Delta\mathbf{h}_c$ as the mean activation difference between attribute-positive
and attribute-negative subsets of CelebA-HQ.

Unlike our perceptual directions where patching toward the degraded concept
is necessary to avoid destructive interference semantic directions encode
high-level structural attributes that are consistent with the mid-trajectory
latent distribution.
Negative patching ($\eta < 0$) is therefore stable here, and we apply our method with positive guidance to steer toward the target attribute.

We report CLIP scores~\citep{radford2021clip}
with the prompts \textit{``a person with a big smile''} and \textit{``a man''},
respectively, to measure how well the steered generations align with the
target concepts.
\begin{figure}[H]
\centering
\setlength{\tabcolsep}{2pt}
\renewcommand{\arraystretch}{1.1}
\captionsetup{skip=2pt}
\resizebox{0.78\columnwidth}{!}{%
\begin{tabular}{@{}>{\centering\arraybackslash}m{1.2em}>{\centering\arraybackslash}m{0.22\columnwidth}>{\centering\arraybackslash}m{0.22\columnwidth}>{\centering\arraybackslash}m{0.22\columnwidth}@{}}
& {\footnotesize Baseline} & {\footnotesize Standard Patching} & {\footnotesize\textbf{Ours}} \\
{\rotatebox[origin=c]{90}{\footnotesize Male}} &
\includegraphics[width=\linewidth]{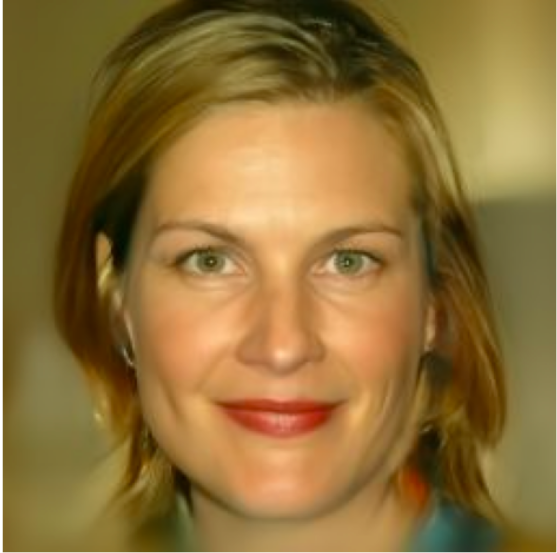} &
\includegraphics[width=\linewidth]{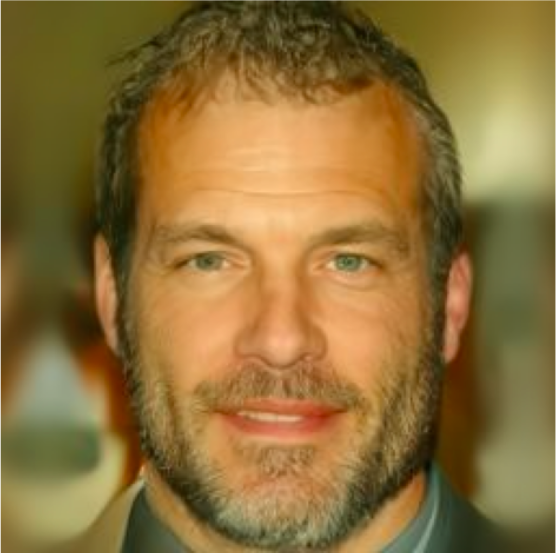} &
\includegraphics[width=\linewidth]{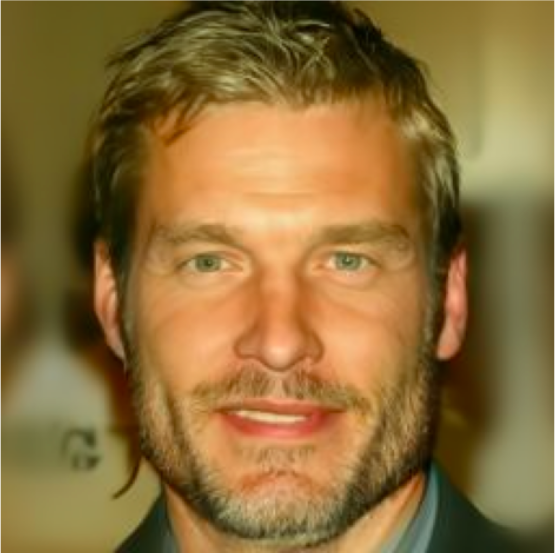}  \\[2pt]
{\rotatebox[origin=c]{90}{\footnotesize Male}} &
\includegraphics[width=\linewidth]{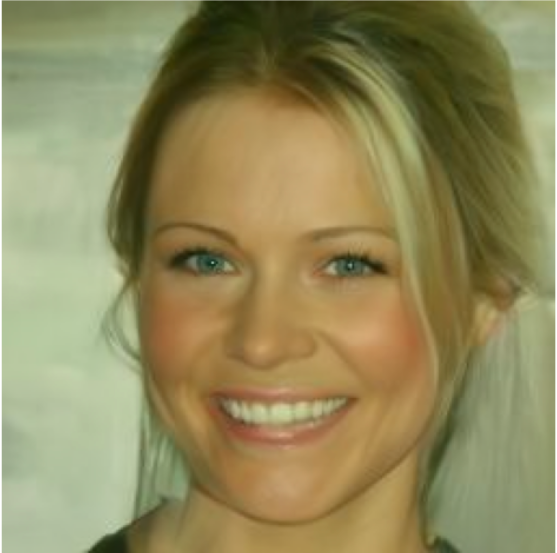} &
\includegraphics[width=\linewidth]{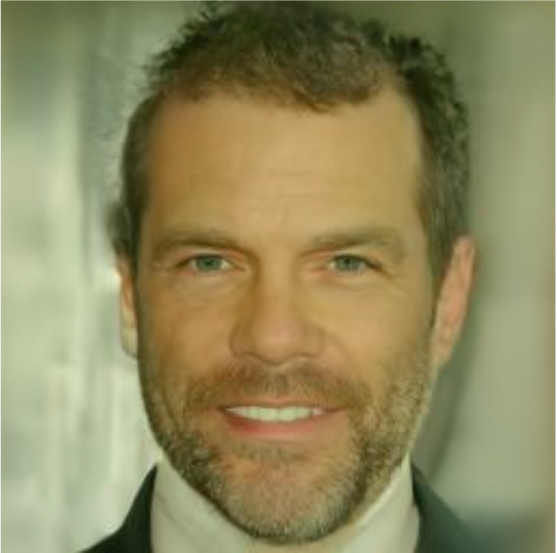} &
\includegraphics[width=\linewidth]{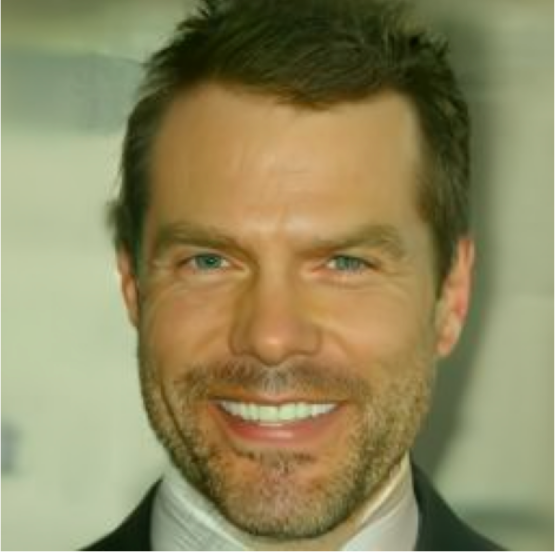} \\[4pt]
{\rotatebox[origin=c]{90}{\footnotesize Smile}} &
\includegraphics[width=\linewidth]{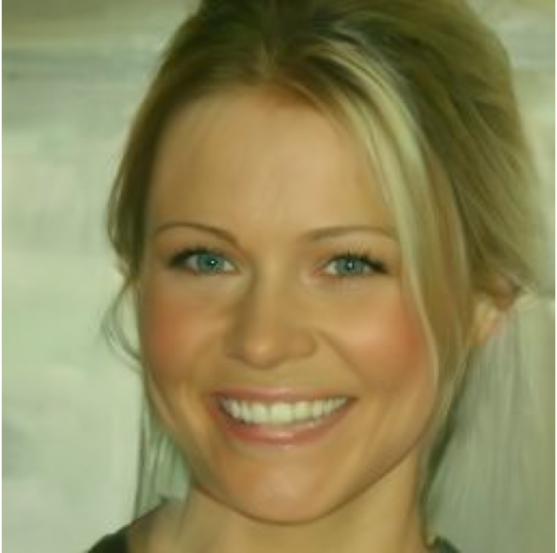} &
\includegraphics[width=\linewidth]{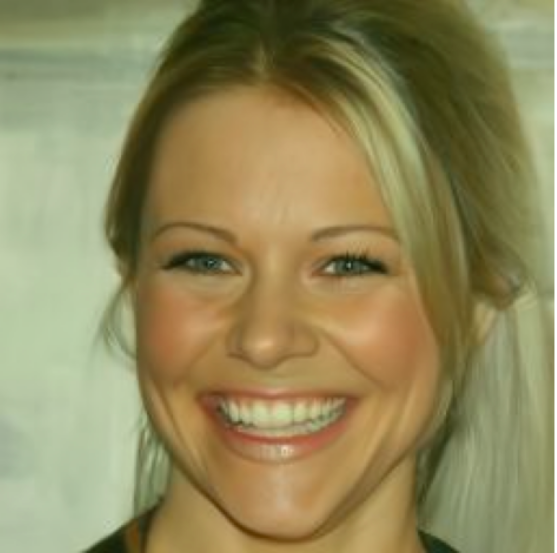} &
\includegraphics[width=\linewidth]{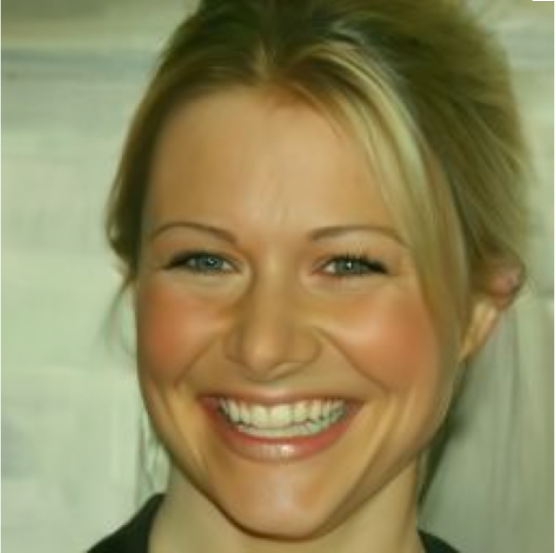} \\[2pt]
{\rotatebox[origin=c]{90}{\footnotesize Smile}} &
\includegraphics[width=\linewidth]{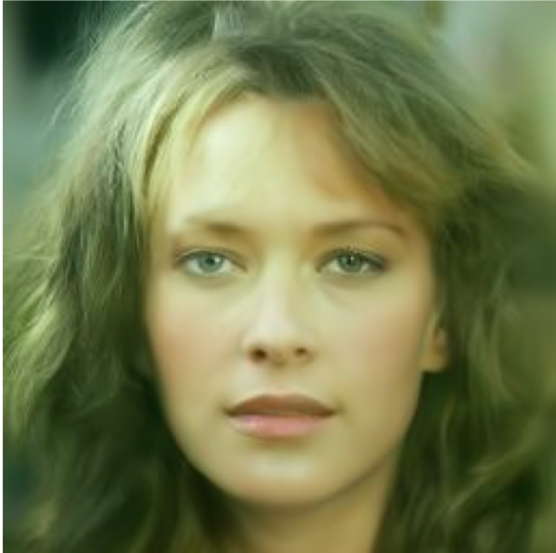} &
\includegraphics[width=\linewidth]{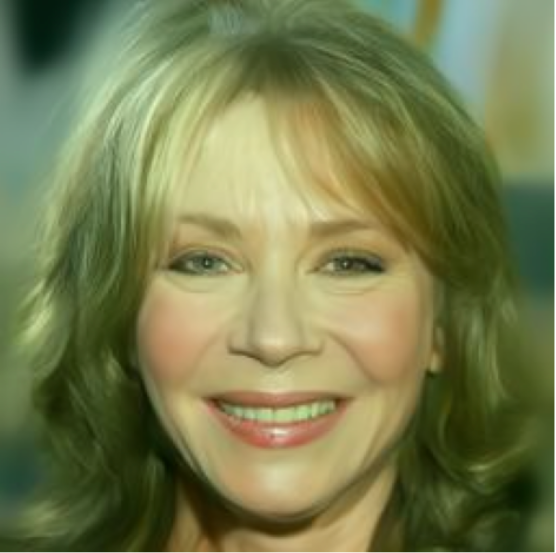} &
\includegraphics[width=\linewidth]{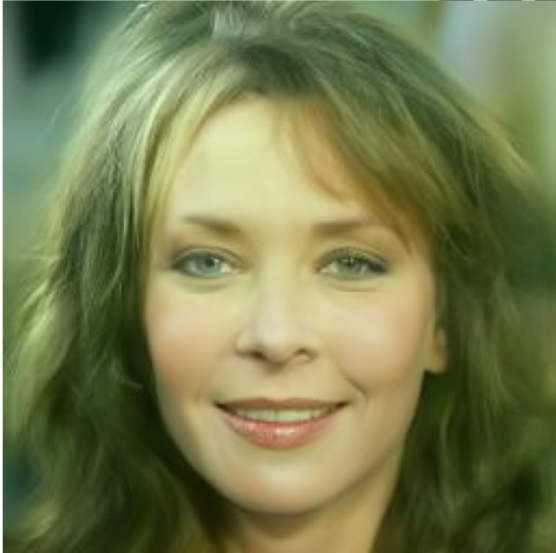} \\
\end{tabular}%
}
\caption{Qualitative comparison between the baseline, Standard Patching and our Method on representative examples with semantically steered directions. This figure shows that our method performs at par, even subjectively better than the patching and it is better at preserving high level features of the image.}
\label{fig:image_grid_4x3}
\vspace{-8pt}
\end{figure}

\begin{table}[ht]
\centering
\small
\caption{Semantic steering on CelebA-HQ evaluated via CLIP score. Higher is better.}
\label{tab:semantic}
\begin{tabular}{lcc}
\toprule
Method & Smiling & Male  \\
\midrule
Baseline       & 0.43 & 0.51 \\
Standard Patching  & \textbf{0.76} & 0.74 \\
Ours          & 0.73 & \textbf{0.75} \\
\bottomrule
\end{tabular}
\end{table}

Our method achieves CLIP alignment comparable to Standard Patching (Table~\ref{tab:semantic}) while more consistently preserving high-level image features (Figure~\ref{fig:image_grid_4x3}).

\subsection{Timestep-Selective Guidance.}
\label{selective_guidance}
Our method applies patched forward pass at every denoising step,
requiring double the compute of standard reverse process.
We propose a \emph{partial guidance} variant: for a trajectory of $T$ steps,
we run the unpatched baseline for the first $(1{-}f)$ fraction of steps
and switch to guidance only for the final $f$ fraction, at a total
cost of $(1{+}f)$ times a baseline run.
This design is motivated by the temporal analysis of linear separability of concepts in  \emph{h-space}  which were quantised with LDA scores in \Cref{apx:vector_sep}, peaking at later denoising timestep, indicating that the
h-space representation becomes linearly separable with respect to perceptual
concepts only as the trajectory approaches the data manifold.

\begin{figure}[ht]
\centering
\setlength{\tabcolsep}{2pt}
\renewcommand{\arraystretch}{1.0}
\captionsetup{skip=2pt}
\begin{tabular}{@{}
  >{\centering\arraybackslash}p{0.152\columnwidth}
  >{\centering\arraybackslash}p{0.152\columnwidth}
  >{\centering\arraybackslash}p{0.152\columnwidth}
  >{\centering\arraybackslash}p{0.152\columnwidth}
  >{\centering\arraybackslash}p{0.152\columnwidth}
  >{\centering\arraybackslash}p{0.190\columnwidth}@{}}
  \includegraphics[width=\linewidth]{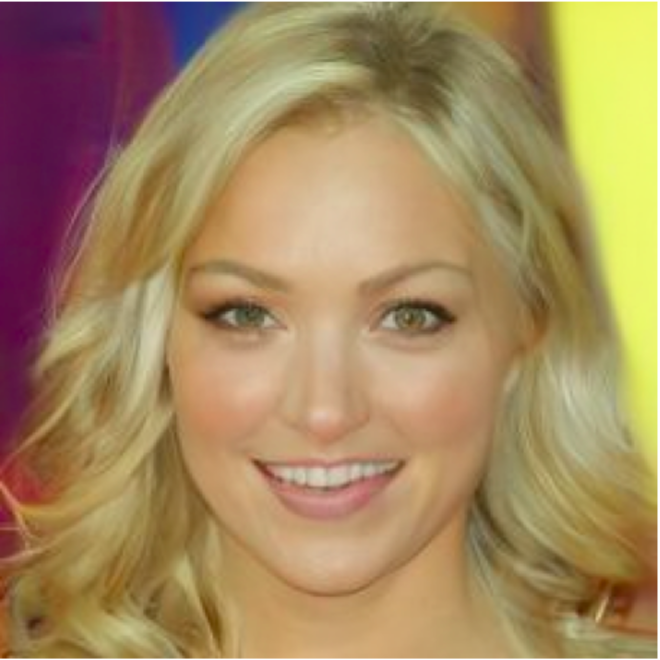} &
  \includegraphics[width=\linewidth]{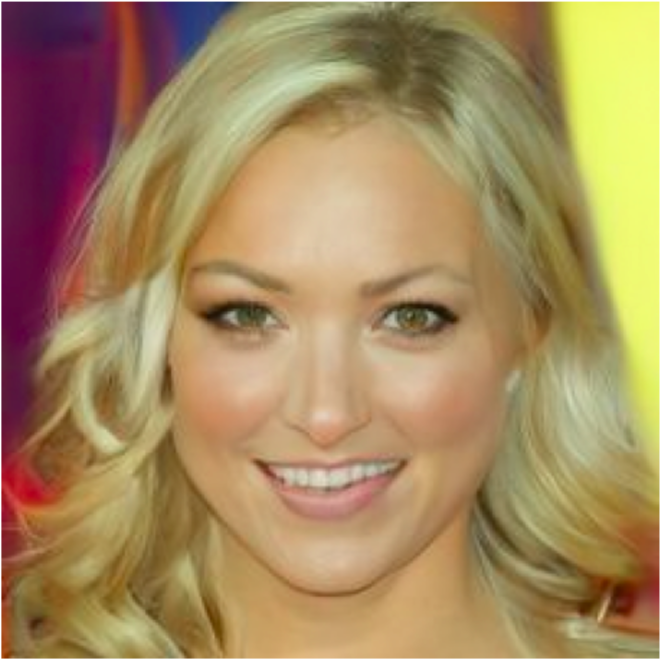} &
  \includegraphics[width=\linewidth]{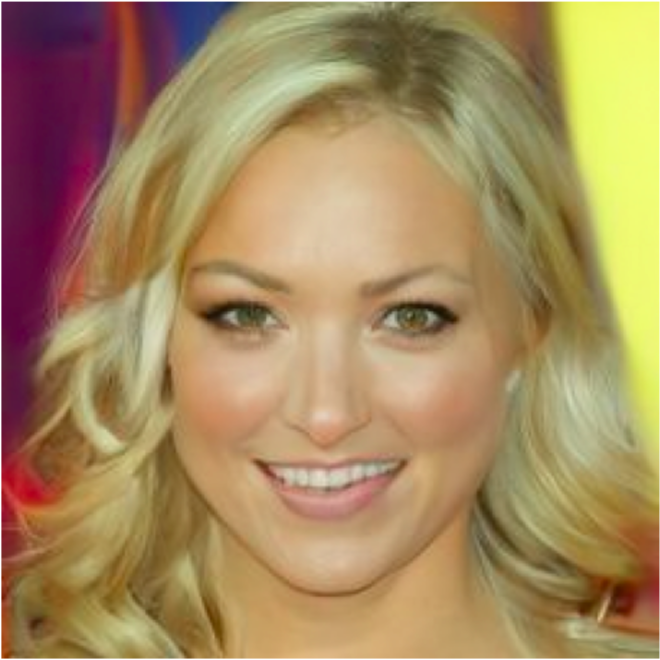} &
  \includegraphics[width=\linewidth]{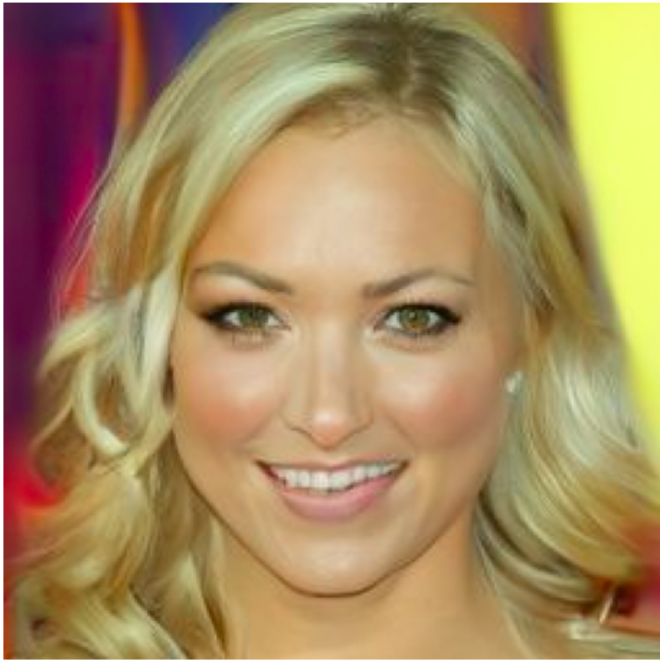} &
  \includegraphics[width=\linewidth]{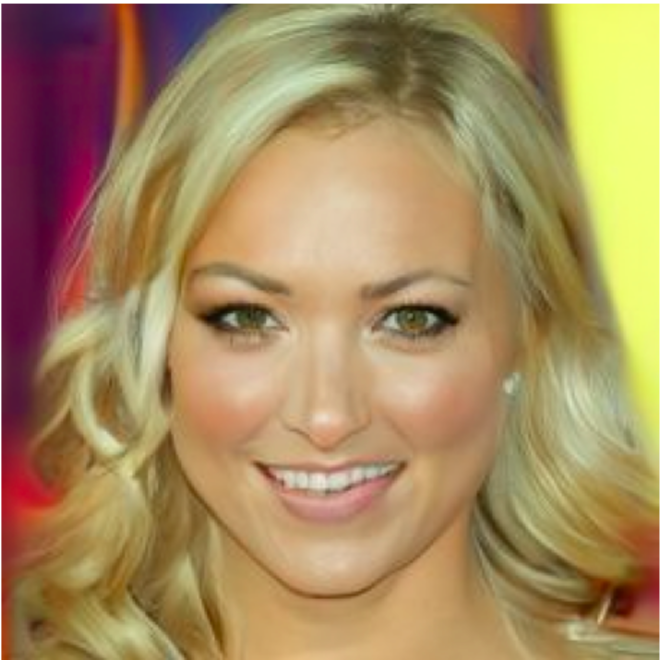} &
  \includegraphics[width=\linewidth]{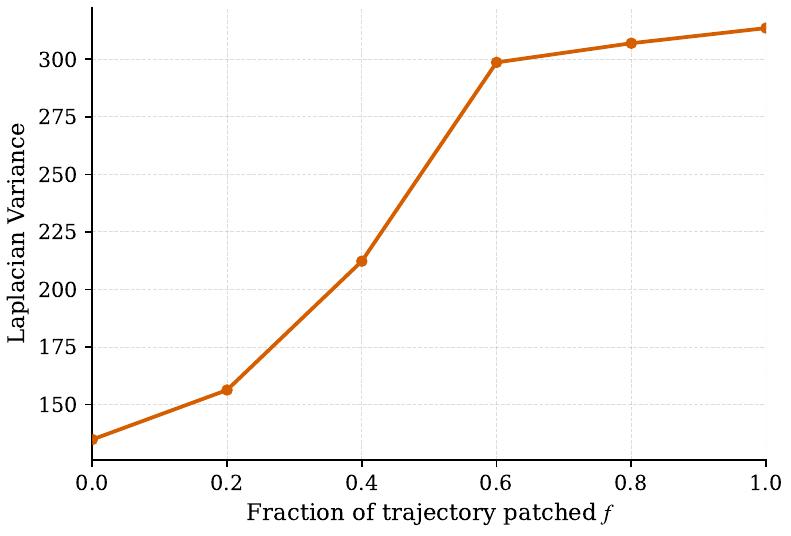} \\[1pt]
  {\scriptsize (a) $f{=}0.2$} &
  {\scriptsize (b) $f{=}0.4$} &
  {\scriptsize (c) $f{=}0.6$} &
  {\scriptsize (d) $f{=}0.8$} &
  {\scriptsize (e) $f{=}1$} &
  {\scriptsize (f) Lap.\ var.\ vs.\ $f$} \\
\end{tabular}
\caption{Effect of guidance fraction $f$. Partial guidance ($f{=}0.6$) recovers ${\sim}90\%$ of full-guidance quality (Laplacian variance) at only $1.6{\times}$ compute vs.\ $2.0{\times}$ for $f{=}1$.}
\label{fig:partial}
\vspace{-4pt}
\end{figure}

Guidance applied before this regime acts on activations where the concept
direction carries little discriminative signal, contributing noise rather
than meaningful steering. \Cref{fig:partial} plots the Laplacian
variance obtained for $f \in \{0.2, 0.4, 0.6, 0.8, 1.0\}$, along with the
corresponding qualitative examples.

These finding confirms that perceptual steering requires intervention only in the late denoising stpes, making our method practical under constrained inference budgets.

\section{Hyperparameters}
\label{apx:hyper}
We publicly release our code for reproducibility \href{https://github.com/dsgiitr/uncond-diffusion-enhancement}{here}. Hyperparameters were tuned manually on a subset of the main dataset. Additionally, details of all hyperparameters have been detailed in the table below.
\begin{table}[H]
\centering
\label{tab:hyperparameters}
\begin{tabular}{lc}
\toprule
\textbf{Hyperparameter} & \textbf{Value} \\
\midrule

\multicolumn{2}{l}{\textit{Model Configuration}} \\[2pt]
\quad Model Name        & \href{https://huggingface.co/google/ddpm-celebahq-256}
 {\texttt{google/ddpm-celebahq-256}} \\
\quad Main Dataset & CelebA-HQ $256\times256$ \\
\quad DDIM Inference Steps   & 30 \\
\quad FID Samples & 10,000\\
\midrule
\multicolumn{2}{l}{\textit{Vector Extraction}} \\[2pt]
\quad Extraction Timestep $(k)$ & 50 \\
\quad Number of Samples $(N)$ & 100\\
\midrule
\multicolumn{2}{l}{\textit{Guidance Scale $(w)$}} \\[2pt]
\quad Our Method & +2.00 \\
\quad Standard Patching   & -1.00 \\
\midrule

\multicolumn{2}{l}{\textit{Patching Scale $(\eta)$}} \\[2pt]
\quad Our Method & 75.00 \\
\quad Standard Patching & -75.00 \\

\bottomrule
\end{tabular}
\end{table}

\end{document}